\documentclass[conference]{IEEEtran}
\IEEEoverridecommandlockouts
\usepackage{cite}
\usepackage{amsmath,amssymb,amsfonts}
\usepackage{algorithmic}
\usepackage{graphicx}
\usepackage{textcomp}
\usepackage{url}
\usepackage{xcolor}
\usepackage{adjustbox}
\usepackage{fancyhdr}

\fancypagestyle{firstpagestyle}{
    \fancyhf{}
    \fancyhead[L]{
    }
    \fancyfoot[L]{
    } 

}

\def\BibTeX{{\rm B\kern-.05em{\sc i\kern-.025em b}\kern-.08em
    T\kern-.1667em\lower.7ex\hbox{E}\kern-.125emX}}
\begin{document}

\title{An Efficient Real Time DDoS Detection Model Using Machine Learning Algorithms}


\author{\IEEEauthorblockN{Debashis Kar Suvra} 
\IEEEauthorblockA{Department of Computer Science and Engineering (CSE)\\ School of Data and Sciences (SDS)\\ 
Brac University\\66 Mohakhali, Dhaka - 1212, Bangladesh}  
\IEEEauthorblockA{debashis.kar.suvra@g.bracu.ac.bd}} 

\maketitle
\thispagestyle{firstpagestyle}

\begin{abstract}

Distributed Denial of Service attacks have become a significant threat to industries and governments leading to substantial financial losses. With the growing reliance on internet services, DDoS attacks can disrupt services by overwhelming servers with false traffic causing downtime and data breaches. Although various detection techniques exist, selecting an effective method remains challenging due to trade-offs between time efficiency and accuracy. This research focuses on developing an efficient real-time DDoS detection system using machine learning algorithms leveraging the UNB CICDDoS2019 dataset including various traffic features. The study aims to classify DDoS and non-DDoS traffic through various ML classifiers including Logistic Regression, K-Nearest Neighbors, Random Forest, Support Vector Machine, Naive Bayes. The dataset is preprocessed through data cleaning, standardization and feature selection techniques using Principal Component Analysis. The research explores the performance of these algorithms in terms of precision, recall and F1-score as well as time complexity to create a reliable system capable of real-time detection and mitigation of DDoS attacks. The findings indicate that RF, AdaBoost and XGBoost outperform other algorithms in accuracy and efficiency, making them ideal candidates for real-time applications.

\end{abstract}


\section{Introduction}

DDoS attack is a common and serious threat to organizations that depend on online services \cite{a11}. DDoS attacks overcome a server or network with excessive traffic rendering it unusable. Often founded by cybercriminals or hacktivists, these attacks aim to disrupt services leading to financial losses, reputation damage and critical infrastructure failures. As DDoS attacks grow more sophisticated detecting and defending against them has become increasingly challenging \cite{b4}. DDoS attackers often target media, universities, online services, financial industries and government servers but small businesses are also at risk \cite{a19}. 

This research develops an efficient real-time DDoS detection system using ML algorithms. Various classifiers are used to classify DDoS and non-DDoS traffic. Preprocessing involves data cleaning, standardization and feature selection with PCA. The study evaluates performance based on precision, recall, F1-score and time complexity with RF, AdaBoost and XGBoost proving most effective for real-time detection and mitigation.

\section{Research Objectives}

The primary objectives of this research: 
\begin{itemize}
    \item Identify the most effective approach with high accuracy and low time complexity.
    \item Create a reliable detection system capable of safeguarding websites by identifying and blocking malicious requests in real-time.
\end{itemize} 



\section{Literature Review}

This paper analyzes recent ML algorithms using the CICDoS2019 dataset for DDoS detection. It finds that GB and XGBoost achieved high accuracies of 99.99\% and 99.98\% respectively with GB also showing a low false alarm rate of around 0.004 \cite{a11}. An ML approach detects DDoS attacks using the NSL-KDD dataset. Algorithms such as RF, KNN and LR are applied and evaluated using cross-validation. RF achieves the highest accuracy of 99.33\% before and after post-pruning. The implications of false positives and negatives are also discussed \cite{b4}. This paper reviews research utilizing algorithms like RF and CNN for DDoS attack detection across various datasets. It discusses methods that split datasets to enhance accuracy and precision or combine mathematical and ML models to improve throughput and resolution time offering a comprehensive analysis of DDoS detection algorithms \cite{a12}. This research identifies the best model for DDoS attack detection using ML algorithms including RF, LR, SVM, KNN and DT. A comparative analysis reveals that the top-performing model achieves a test accuracy of 99.95\% \cite{b1}. 

This research proposes an enhanced method for detecting these attacks by incorporating ML techniques \cite{a13}. This study creates a DDoS dataset and classifies attacks through behavioral analysis using a virtual lab to capture both attack and benign traffic. Various ML techniques are applied with KNN outperforming others. The unique heterogeneous DDoS dataset results in RF and KNN achieving the highest classification accuracies of 99.44\% and 99.58\% respectively \cite{b3}. The CICDDoS2019 dataset is used to develop ML models for DDoS attack detection. A balanced dataset with 360,000 records and 15 key features is created using random sampling and feature engineering. Various classifiers including DT, RF, NB, Stochastic GB and KNN are trained and tested. RF achieves the highest accuracy of over 99\% in both datasets demonstrating the effectiveness of ML algorithms in real-time DDoS detection \cite{a14}. ML algorithms classify normal and DDoS attack traffic using the CICDDoS2019 dataset from the Canadian Institute of Cyber Security. Among the four classification methods tested including KNN, DT RF and ANN. The ANN achieves the highest accuracy in DDoS detection \cite{b2}.

Najar and Naik apply various ML techniques to detect DDoS attacks using classifiers like RF, KNN, SVM and DT on the UNSW-NB15 benchmark dataset. The RF model outperforms the others achieving 94.14\% accuracy on training data and 91.88\% on the test dataset, while also showing a higher detection rate and lower false alarm rate for distinguishing malicious from benign traffic \cite{a16}. A new approach for detecting DDoS attacks analyzes traffic flow traces and generates a confusion matrix. NB and RF are used to classify traffic as normal or abnormal based on existing datasets. The results show that the NB algorithm outperforms the RF algorithm in this classification task \cite{a15}.

In this research, we develop an efficient real-time DDoS detection system using ML algorithms on the UNB CICDDoS2019 dataset. It classifies DDoS and non-DDoS traffic using various classifiers with preprocessing techniques. The findings show that RF, AdaBoost and XGBoost outperform other algorithms in accuracy and efficiency making them suitable for real-time applications.

\section{Dataset Analysis}

\subsection{Dataset}
We are using a dataset from the Canadian Institute for Cybersecurity (CIC) which includes 87 traffic features. Each holds its significance and value. 

\subsection{Dataset Feature Description}

We use the CIC 2019 DDoS Evaluation Dataset for training which includes 87 traffic features each with distinct importance. Key features involve source and destination ports, flow duration and the number of packets and bytes per second in both forward and backward directions. The dataset also tracks packet lengths (maximum, minimum, mean and standard deviation), flow inter-arrival times, push and URG flags, header lengths, packet rates, download/upload ratios, bulk rates, sub-flows and active time before becoming idle. These features provide a detailed analysis of network traffic.

\section{Methodology}

\begin{figure}[ht]
\centering
\includegraphics[scale=0.33]{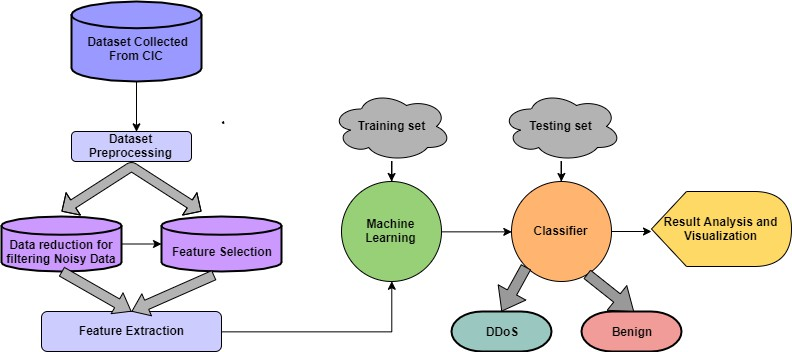}
\caption{Workflow Diagram of Prediction model}
\label{fig: b}
\end{figure} 

Figure-\ref{fig: b} summarizes the approach used to identify DDoS and non-DDoS traffic. We begin by selecting a large dataset from the CIC 2019 DDoS dataset. The data is preprocessed by removing any infinite values and noisy data. A standard scaler is applied to standardize the dataset. Feature selection is then carried out using WEKA employing several techniques discussed in the feature selection section. Finally, the preprocessed data is provided into various ML classifiers and the results are analyzed in the result analysis section.

\subsection{Data preprocessing}

The process can be divided into several steps:

\begin{itemize}
    \item \textbf{Removing Noisy and Missing Values:} Initially, we eliminate noisy and missing data as raw datasets often contain such issues. Noisy data refers to irrelevant information and our dataset also had infinity or NaN values which we resolve.
    \item \textbf{Removing Duplicate Values:} There are a few duplicate entries in the raw dataset that we removed.
    \item \textbf{Data Standardization:} This involves modifying the data to ensure consistency. We used the StandardScaler from sklearn to standardize the dataset, setting a specific range for the data.
\end{itemize}

\begin{figure}[ht]
\centering
\includegraphics[scale=0.45]{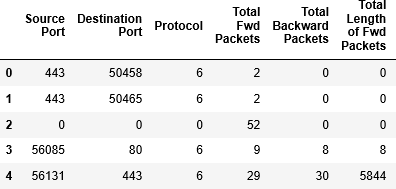}
\caption{Before data preprocessing with StandardScaler}
\label{fig: bb}
\end{figure} 

Figure-\ref{fig: bb} shows network traffic data with key features. The Source Port and Destination Port columns indicate the port numbers of the source and destination devices. The Protocol column identifies the communication protocol where '6' represents TCP. The Total Fwd Packets and Total Backward Packets columns denote the number of packets sent in forward and backward directions while the Total Length of Fwd Packets column records the total length of all forward packets in each flow.
In the Figure-\ref{fig: bb}:
\begin{itemize}
    \item Row 0 shows traffic from Source Port 443 to Destination Port 50458 using protocol 6, with 2 forward and 0 backward packets.
    \item Row 4 has traffic from Source Port 56131 to Destination Port 443, with 29 forward packets, 30 backward packets and a total forward packet length of 5844 bytes.
\end{itemize}

\begin{figure}[ht]
\centering
\includegraphics[scale=0.45]{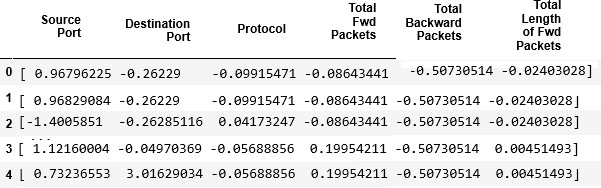}
\caption{After data preprocessing with StandardScaler}
\label{fig: bbb}
\end{figure} 

Figure-\ref{fig: bbb} presents a normalized dataset with standardized feature values. Each feature is scaled to have a mean near zero to enhance the performance of ML models. Figure-\ref{fig: bbb} shows that row 0 has the normalized values:
\begin{itemize}
    \item Source Port: 0.96796225, Destination Port: -0.26229, Protocol: -0.09915471.
\end{itemize}

This normalization process helps in reducing the impact of the different scales of each feature making the dataset suitable for ML algorithms.

\subsection{Feature Selection}

We use WEKA to process our dataset. WEKA is recognized as one of the leading software tools for data mining equipped with various ML algorithms. Our dataset is quite large, initially containing 84 features and approximately 1 million records with 40\% of the data representing benign traffic and the rest consisting of DDoS attack data. The focus is on identifying the most relevant features for detecting various DDoS attacks including UDP flood, NTP amplification, SYN flood, HTTP flood and ICMP (ping) flood while keeping the dataset manageable.

\begin{figure}[ht]
\centering
\includegraphics[scale=0.12]{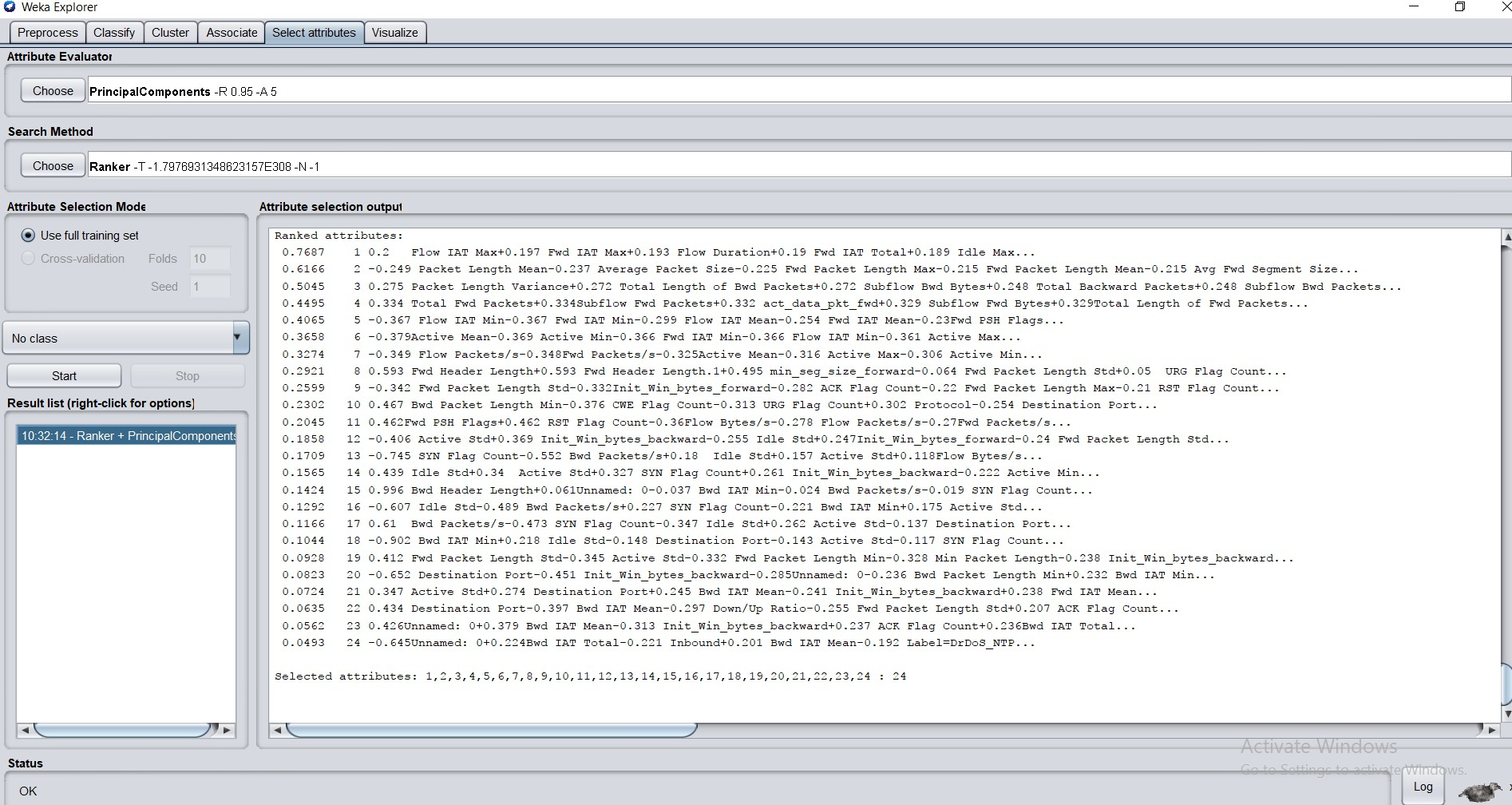}
\caption{PCA for selecting features using WEKA}
\label{fig: c}
\end{figure}

Initially, we apply a filter for preprocessing. But this resulted in only 6 features which is insufficient. After obtaining 11 features, we apply PCA which generates 24 key features. Using WEKA's "Attribute Evaluator" and Ranker search method, PCA consistently provides the best results. We also explore other methods like the wrapper method and attribute selection with NB and LR using ten-fold cross-validation. However, PCA consistently provides the best results. So we chose the features identified by PCA. Figure-\ref{fig: c} shows the PCA for selecting features.

\subsection{Selected Features}

\begin{figure}[ht]
\centering
\includegraphics[scale=0.22]{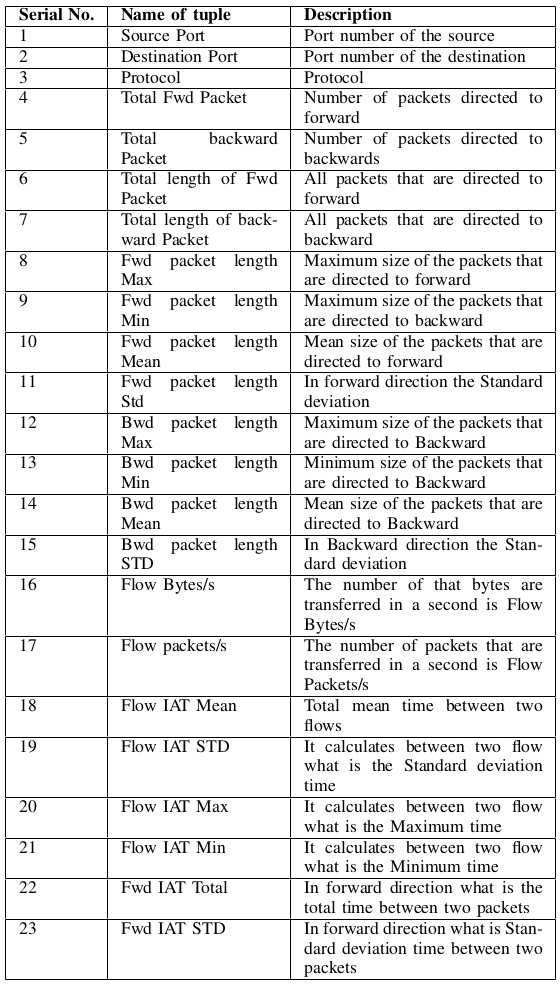}
\caption{Selected Features Description}
\label{fig: cnn}
\end{figure}

Figure-\ref{fig: cnn} presents a description of the selected features. Our dataset is primarily a customized version created by combining benign data with DDoS attack types such as NTP, DNS and UDP from the original dataset. This resulting dataset is imbalanced.

\subsection{Trainning or Testing machine learning classifier}

After feature selection, the dataset is split into a training set (60\%) and a testing set (40\%). Various ML classifiers are trained using the training set and the testing set is used to evaluate the model's ability to detect new attacks and provide accurate results.

\subsection{Data visualization}

Data visualization graphically represents data to indicate relationships between variables. Using Python libraries like Matplotlib, Seaborn and Scikit Plot, we generate various plots.

\subsubsection{Heat Map}

\begin{figure}[ht]
\centering
\includegraphics[scale=0.45]{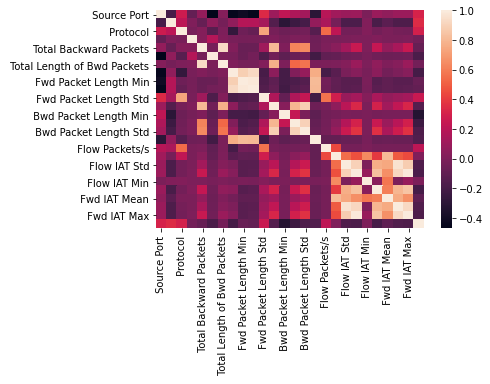}
\caption{Heat Map}
\label{fig: ccn}
\end{figure}

We use Python's Seaborn library to create a heatmap of the dataset for visualization. The heatmap is generated with the characteristic variables as both column and row headers. Figure-\ref{fig: ccn} shows a visual representation of the correlations between the variables in a two-dimensional space.

Figure-\ref{fig: ccn} shows a heat map that visualizes the relationships between features in the dataset. In the heat map, color intensity indicates correlation strength, darker shades show weaker correlations while lighter shades represent stronger ones. A diagonal line of brighter squares from top left to bottom right indicates each feature's perfect correlation with itself. Lighter areas away from the diagonal highlight stronger correlations between features, while darker areas show weaker or no correlations. This visualization helps understand feature interdependencies which are important for model performance and feature selection in ML.

\subsubsection{Correlation Matrix}

\begin{figure}[ht]
\centering
\includegraphics[scale=0.35]{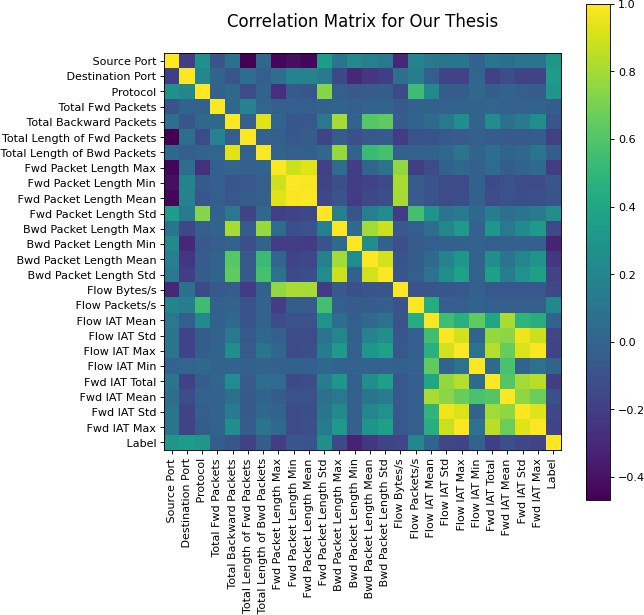}
\caption{Correlation Matrix}
\label{fig: cc}
\end{figure}

The matrix is symmetrical with values in the lower left corresponding to those in the upper right. Diagonally the upper left and lower right points represent the relationships between the attributes as shown in Figure-\ref{fig: cc}.

Figure-\ref{fig: cc} shows a correlation matrix that visualizes relationships between various network traffic and packet-related features. The color scale from dark purple (strong negative correlation) to bright yellow (strong positive correlation) represents the strength and direction of correlations between approximately 25 variables including Source Port, Destination Port, Protocol, packet counts, packet lengths, flow bytes and Inter-Arrival Time metrics. The symmetric matrix reveals strong correlations among IAT-related metrics and between packet counts and their respective lengths. Teal and dark blue cells indicate weak or no correlation. This visualization provides key insights into feature interconnections useful for network behavior analysis, pattern detection and feature selection in ML models for security and performance optimization.

\subsubsection{Count Plot}

\begin{figure}[ht]
\centering
\includegraphics[scale=0.48]{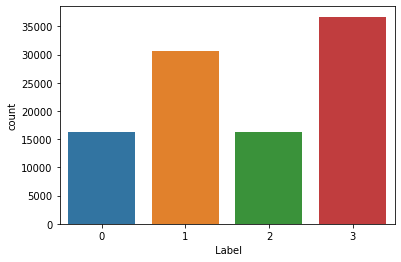}
\caption{Count Plot}
\label{fig: ccb}
\end{figure}

We use Python to create a count plot for the "Label" category which includes 4 types: 'BENIGN', 'DDoS-DNS', 'DDoS-UDP' and 'DDoS-NTP'. The count plot shows the frequency of each label providing a quick visual overview of how often each item appears, as shown in Figure-\ref{fig: ccb}.

Figure-\ref{fig: ccb} shows a bar chart demonstrating the data distribution across 4 categories labeled 0 to 3 on the x-axis with the y-axis indicating counts up to around 35,000. Each category is depicted with a different colored bar: blue for category 0, orange for category 1, green for category 2 and red for category 3. The heights of the bars represent the frequency for each category with category 3 having the highest count (around 35,000) followed by category 1 (about 30,000) and categories 0 and 2, each with lower counts of approximately 15000-17000. This chart effectively visualizes the relative proportions of these four categories within the dataset showing a clear dominance of categories 3 and 1 over categories 0 and 2.

\section{Result Analysis and Real Time Implementation}

\subsection{Result Analysis}

\begin{table}[!ht]
    \centering
    \caption{Performances of various Algorithms}
    \begin{tabular}{l l l l l}
    \hline
        Algorithm & 0 & 1 & 2 & 3  \\ 
         & Pre. Re. F1. & Pre. Re. F1. & Pre. Re. F1. & Pre. Re. F1. \\ \hline
        LR & .87   .97   .92 & .85 .92 .88 & .84 .57 .68 & 0.98 1.0 .99 \\ \hline
        KNN  & .99 .99 .99 & .95 .97 .96 & .94 .89 .91 & 1.00 1.0 1.0 \\ \hline
        RF  & .99 1.0 .99 & .99 .98 .98 & .96 .97 .96 & 1.00 1.0 1.0 \\ \hline
        AdaBoost  & .99 1.0 .99 & .99 .98 .98 & .96 .97 .96 & 1.00 1.0 1.0\\ \hline
        SVM  & .89 .99 .94 & .89 .93 .91 & .87 .67 .76 & 0.99 1.0 .99  \\ \hline
        DT  & .99 .99 .99 &.98 .98 .98 & .96 .95 .96 & 1.00 1.0 1.0 \\ \hline
        XGBoost  & .98 1.0 .99 & .99 .97 .98 & .95 .96 .95 & 1.00 1.0 1.0\\ \hline
        NB  & .90 .64 .75 & .74 .94 .83 & .31 .17 .22 & 0.89 .99 .94\\ \hline
    \end{tabular}
    \label{tab:datasettt}
\end{table}

Table-\ref{tab:datasettt} presents the performance of various mML algorithms based on precision (Pre.), recall (Re.), and F1-score (F1.) for four different classes: 0 (Benign), 1 (DDoS-DNS), 2 (DDoS-NTP), and 3 (DDoS-UDP). The performance metrics indicate the algorithms' effectiveness with RF, AdaBoost and XGBoost achieving the highest precision, recall, and F1 scores across all classes. At the same time, NB shows lower performance, especially for DDoS-NTP.

\begin{table}[htbp]
\caption{Accuracy, AUC Score, Average Execution Time of Algorithms}{
\begin{tabular}{|p{1.1cm}|p{0.9cm}|p{1.3cm}|p{3.7cm}|}
\hline
Algorithms & Accuracy & AUC Score & Execution Time in second (Avg) \\
\hline
LR & 0.9174 & 0.9335 & 5.556 s \\
\hline
NB & 0.7733 & 0.889 & 5.189 s \\
\hline
KNN & 0.9688 & 0.996 & 5.045 s \\
\hline
DT & 0.9850 & 0.997 & 2.015 s\\
\hline
RF & 0.9880 & 0.998 & 9.71 s \\
\hline
Adaboost & 0.9876 & 0.997 & 96.122 s \\
\hline
SVM & 0.9212 & 0.995 & 69.301 s \\
\hline
XGboost & 0.9824 & 0.997 & 24.037 s \\
\hline
\end{tabular} }
\label{tab: ab}
\end{table}

Table-\ref{tab: ab} summarizes the performance of different ML algorithms in terms of accuracy, AUC score and average execution time. RF and DT exhibit the highest accuracy and AUC scores, while KNN performs well with low execution time. On the other hand, NB shows lower accuracy and AUC scores. AdaBoost though highly accurate has the longest execution time among the algorithms.

\begin{figure}[ht]
\centering
\includegraphics[scale=0.2]{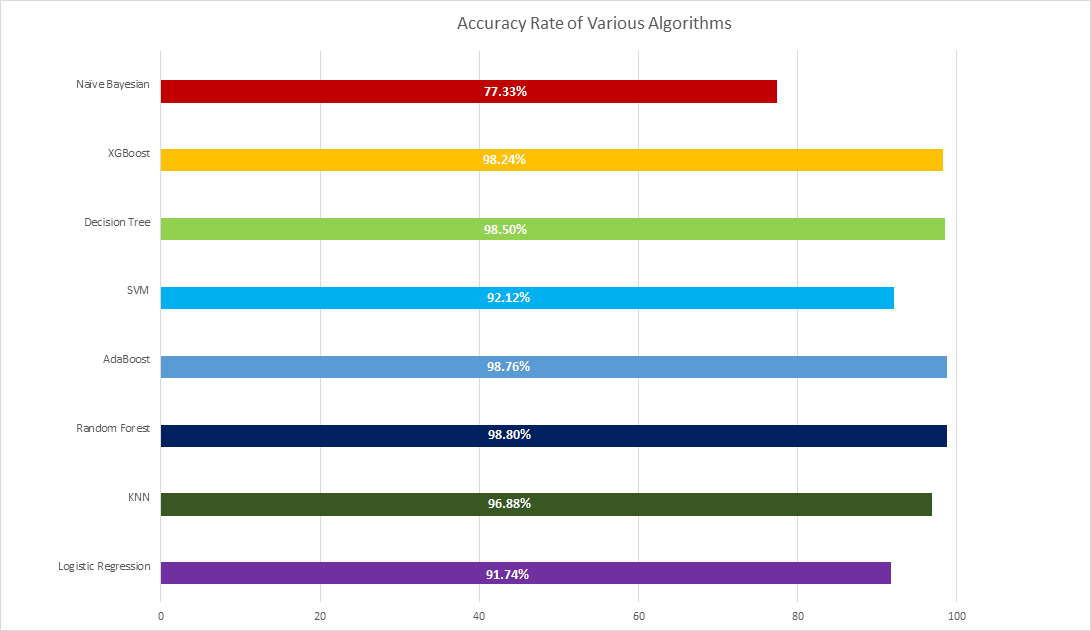}
\caption{Accuracy rate of Various Algorithms}
\label{fig: f}
\end{figure} 

Figure-\ref{fig: f} compares the accuracy rates of eight different ML algorithms. Each algorithm is represented by a bar of a different color with the accuracy percentage labeled on each bar. The x-axis ranges from 0 to 100 representing the accuracy percentage. RF offers the highest accuracy at 98.80\%, closely followed by AdaBoost and DT, while NB has the lowest accuracy at 77.33\%. This visualization effectively represents the relative performance of these algorithms in terms of accuracy allowing for quick comparison across different ML techniques.

\begin{figure}[ht]
\centering
\includegraphics[scale=0.38]{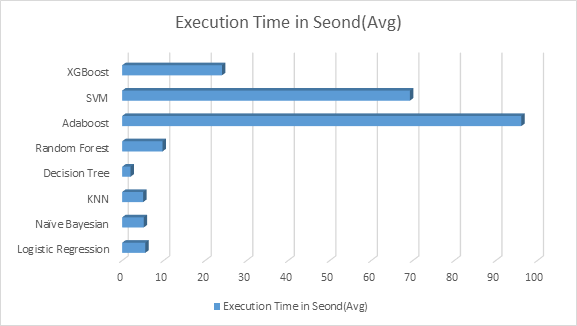}
\caption{Execution Time of various Algorithms}
\label{fig: g}
\end{figure} 

Figure-\ref{fig: g} presents a horizontal bar chart that compares the average execution times of eight different ML algorithms. The x-axis shows execution times in seconds ranging from 0 to 100. AdaBoost has the longest execution time nearly reaching 100 seconds followed by SVM at around 60 seconds and XGBoost at approximately 25 seconds. RF requires about 10 seconds, while DT, KNN, NB and LR have very short execution times represented by small bars that are barely visible. This chart emphasizes the significant differences in computational efficiency among the algorithms with AdaBoost and SVM being the slowest.

\subsubsection{Confusion Matrix}

\begin{figure}[ht]
\centering
\includegraphics[scale=0.13]{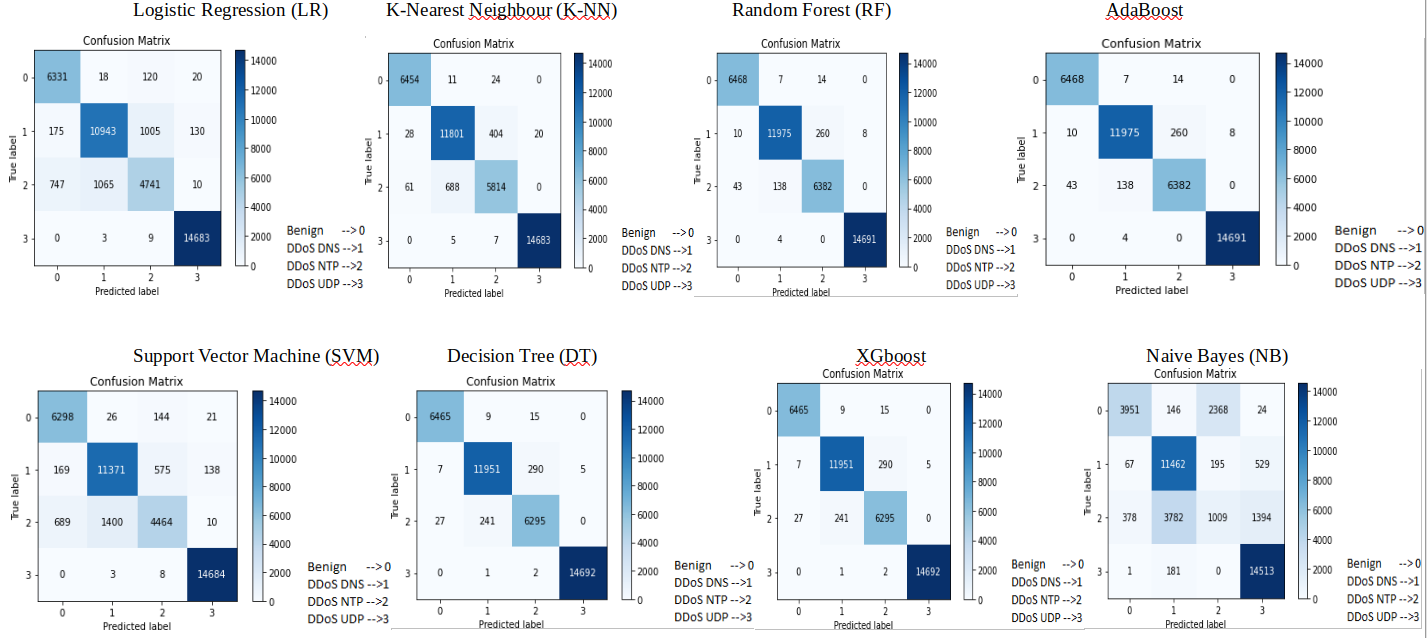}
\caption{Confusion matrices of various Algorithms}
\label{fig: d}
\end{figure}

Figure-\ref{fig: d} shows a series of confusion matrices for various ML classifiers used in DDoS attack detection. Each matrix has true labels on the y-axis and predicted labels on the x-axis with 4 categories: 'Benign', 'DDoS-DNS', 'DDoS-NTP' and 'DDoS-UDP' corresponding to numerical values 0, 1, 2 and 3. Diagonal elements indicate correctly classified instances for each attack type while off-diagonal elements show misclassifications. Color intensity reflects prediction frequency with darker shades indicating higher values. Most classifiers demonstrate high accuracy in identifying 'Benign' and 'DDoS-UDP' traffic although some variations exist in distinguishing between the DDoS attack types.

\subsubsection{ROC Curve}

\begin{figure}[ht]
\centering
\includegraphics[scale=0.13]{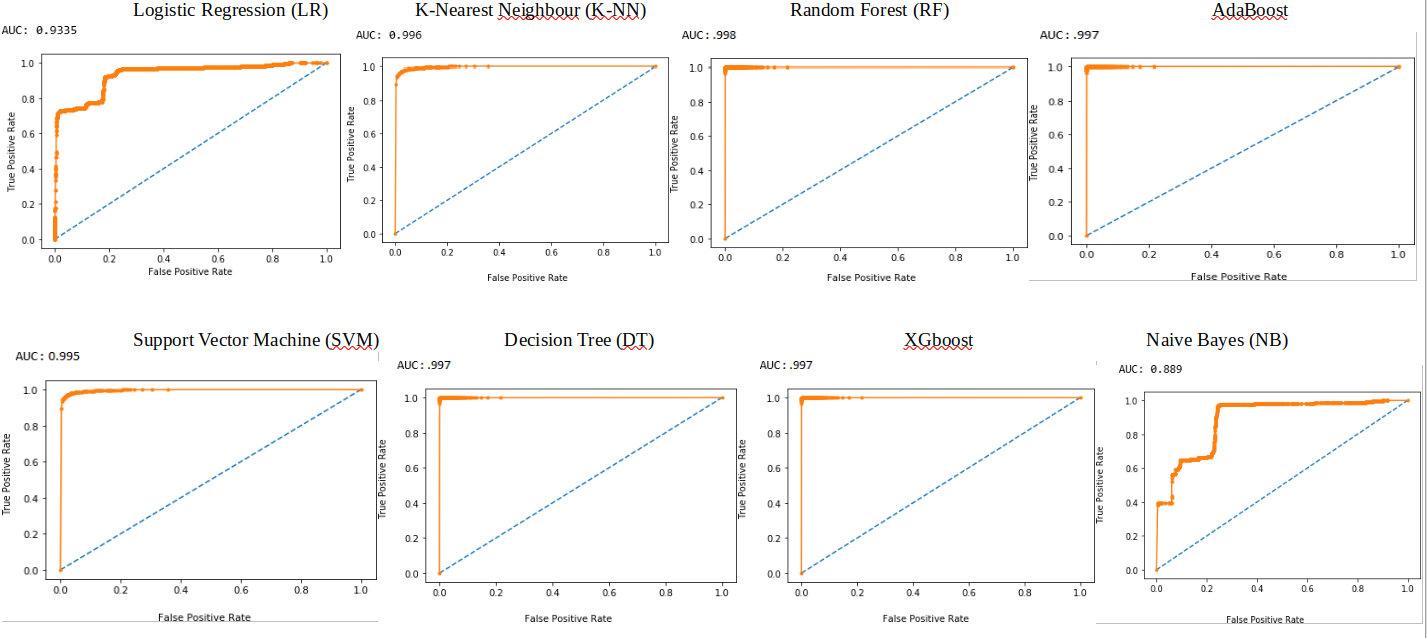}
\caption{ROC Curve of various Algorithms}
\label{fig: e}
\end{figure} 

Figure-\ref{fig: e} presents a comparison of Receiver Operating Characteristic curves for 8 different ML models. Each subplot shows the ROC curve for a specific model with the False Positive Rate on the x-axis and the True Positive Rate on the y-axis. A diagonal blue dashed line indicates the performance of a random classifier while the Area Under the Curve score reflects each model's overall performance. Most models achieve excellent results with AUC scores exceeding 0.99 except for LR (0.9335) and NB (0.889) which perform slightly lower. The ROC curves for high-performing models (KNN, RF, AdaBoost, SVM, DT and XGBoost) are nearly rectangular signifying near-perfect classification. This visualization provides a quick way to compare the models' discriminative abilities at different classification thresholds.

\subsection{Result Comparison}

\begin{table}[!ht]
    \centering
    \caption{Model performance comparison with previous studies}
    \begin{tabular}{l l l l l l l}
    \hline
        Model & Accuracy & Precision & Recall & F1- & AUC & Execution \\
        Name &  &  &  & Score & Score  & Time(s) \\ \hline
        DT \cite{a14} & 0.992 & 0.992 & 0.992 & 0.992 & - & -\\ 
        RF \cite{a14} & 0.994 & 0.994 & 0.994 & 0.994 & -  & -\\
        KNN \cite{a14} & 0.925 & 0.925 & 0.925 & 0.925 & - & -\\
        NB \cite{a14} & 0.493 & 0.493 & 0.493 & 0.493 & - & - \\ \hline
        NB \cite{a15} &  0.9090 \\
        RF \cite{a15} &  0.7817 \\ \hline
        RF \cite{a16} & 0.9188 & - & - & - & - & - \\ \hline
        KNN \cite{a17} & 0.983 & - & - & - & - & - \\ 
        RF \cite{a17} & - & - & - & - & - & - \\ 
        DT \cite{a17} & - & - & - & - & - & - \\ 
        NB \cite{a17} & 0.9448 & - & - & - & - & - \\  \hline
        KNN \cite{a18}  &  0.95 & 0.66 & - & - & 63 & -\\ 
        RF \cite{a18} &  0.99 & 0.99 & - & - & 88 & -\\
        NB \cite{a18} &  0.26 & 0.66 & - & - & 188 & -\\ \hline
        LR & 0.9174 & 0.98 & 1.00 & 0.99 & 0.9335 & 5.556 \\
        NB & 0.7733 & 0.89 & 0.99 & 0.94 & 0.889 & 5.189  \\
        KNN & 0.9688 & 1.00 & 1.00 & 1.00 & 0.996 & 5.045  \\
        DT & 0.9850 & 1.00 & 1.00 & 1.00 & 0.997 & 2.015 \\
        RF & 0.9880 & 1.00 & 1.00 & 1.00 & 0.998 & 9.71  \\
        Adaboost & 0.9876 & 1.00 & 1.00 & 1.00 & 0.997 & 96.122  \\
        SVM & 0.9212 & 0.99 & 1.00 & 0.99 & 0.995 & 69.301  \\
        XGboost & 0.9824 & 1.00 & 1.00 & 1.00 & 0.997 & 24.037 \\
        \hline
    \end{tabular}
    \label{tab:daa}
\end{table}

Table \ref{tab:daa} compares various machine learning models for DDoS attack detection. Our model excels due to its high accuracy with the DT and RF achieving 98.50\% and 98.80\% respectively, while maintaining perfect precision, recall and F1 scores. The DT model's execution time of only 2.015 seconds makes it suitable for real-time applications. Additionally, we employ Principal Component Analysis for effective feature selection which enhances model performance without unnecessary complexity. The RF model benefits from ensemble learning improving prediction accuracy and reducing overfitting. Overall, our approach outperforms previous studies positioning our model as a reliable and efficient tool for DDoS detection and network security.

\subsection{Real Time Implementation}

\begin{figure}[ht]
\centering
\includegraphics[scale=0.35]{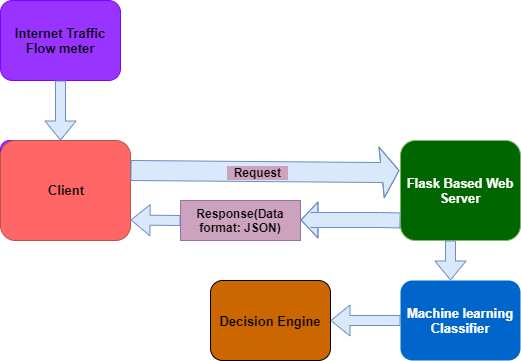}
\caption{Client-Server workflow model of Real-time implementation}
\label{fig: h}
\end{figure} 

Figure-\ref{fig: h} describes a simplified system architecture for an internet traffic flow analysis system. It consists of five main components connected by directional arrows. At the top is an "Internet Traffic Flow meter" (purple) which feeds data to a "Client" (red). The client sends a "Request" to a "Flask Based Web Server" (green). The web server is connected to a "Machine learning Classifier" (blue), which in turn feeds into a "Decision Engine" (brown). The web server responds to the client with "Response(Data format: JSON)". This diagram visualizes the flow of data and decisions in the system from traffic measurement through processing and analysis to the final decision-making stage. Using a Flask-based server and JSON data format suggests a modern, web-based implementation of this traffic analysis and classification system.

\begin{figure}[ht]
\centering
\includegraphics[scale=0.23]{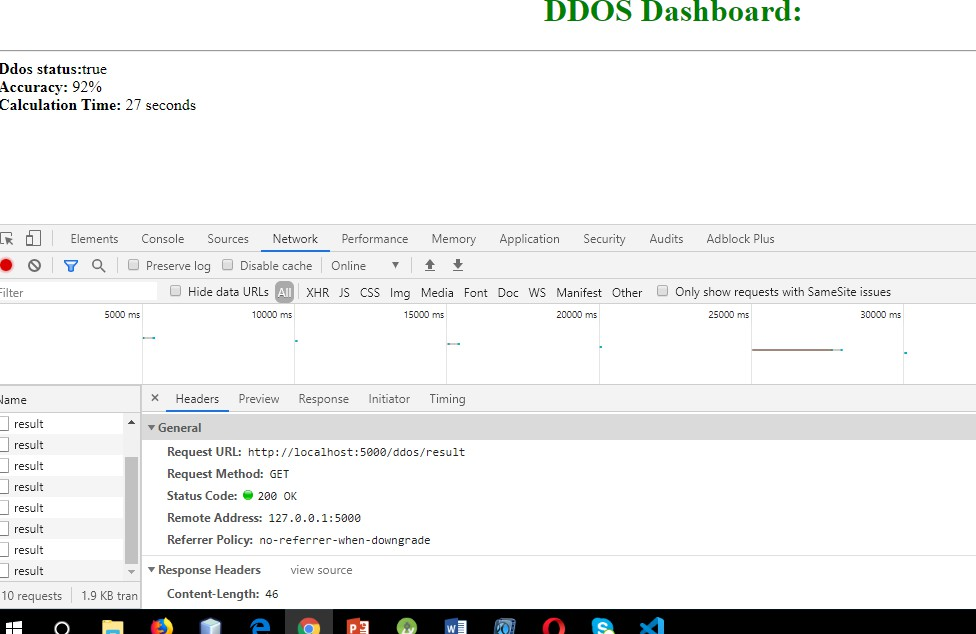}
\caption{Real Time implementation Dashboard}
\label{fig: i}
\end{figure} 

Figure-\ref{fig: i} shows a DDoS Dashboard interface feasible from a web browser's developer tools. At the top, it shows key metrics including a "true" DDoS status (indicating an attack is detected), 92\% accuracy and a calculation time of 27 seconds. Below a network activity timeline presents requests over time. The lower part of the image provides details of a specific request including the request URL (http://localhost:5000/ddos/result), method (GET), status code (200 OK) and other HTTP headers. This dashboard serves as a real-time monitoring tool for analyzing potential DDoS attacks offering both overview statistics and detailed request information for further analysis.

\section{Conclusion}

This study successfully presents an efficient DDoS detection system employing ML techniques validated on the CICDDoS2019 dataset. Through preprocessing, feature selection and the implementation of multiple classifiers, the research emphasizes RF, AdaBoost and XGBoost as the most effective algorithms achieving high accuracy and low execution times. The feature selection process using PCA significantly enhances model performance making the system suitable for real-time DDoS attack detection. For future work, we will focus on expanding the system's capabilities by integrating deep learning techniques and transitioning to a real-time web server and database infrastructure. Additionally, we aim to develop a hybrid classifier that delivers greater accuracy with reduced computational time compared to existing models. Overall, our proposed model is designed to encounter security demands effectively while improving detection speed and accuracy.

\bibliographystyle{IEEEtran}

\end{document}